\def\year{2021}\relax
\documentclass[letterpaper]{article} 
\usepackage{aaai21}  
\usepackage{times}  
\usepackage{helvet} 
\usepackage{courier}  
\usepackage[hyphens]{url}  
\usepackage{graphicx} 
\usepackage{natbib}  
\usepackage{caption} 
\usepackage{amsmath}
\usepackage{amsfonts}
\usepackage{subcaption}
\usepackage{makecell}
\usepackage{xcolor}
\newdimen\figrasterwd
\figrasterwd\textwidth
\title{Domain Adaptation In Reinforcement Learning Via Latent Unified State Representation}
\author{

    Jinwei Xing\textsuperscript{\rm 1}, Takashi Nagata\textsuperscript{\rm 2}, Kexin Chen\textsuperscript{\rm 1}, Xinyun Zou\textsuperscript{\rm 2}, Emre Neftci \textsuperscript{\rm 1, \rm 2} , Jeffrey L. Krichmar\textsuperscript{\rm 1, \rm 2}\\
}
\affiliations{

    \textsuperscript{\rm 1}Department of Cognitive Sciences\\
    \textsuperscript{\rm 2}Department of Computer Science\\

  University of California,Irvine, USA\\
    \{jinweix1,takashin,kexinc3,xinyunz5,eneftci,jkrichma\}@uci.edu 

}

\begin{document}

\maketitle

\begin{abstract}
Despite the recent success of deep reinforcement learning (RL), domain adaptation remains an open problem. Although the generalization ability of RL agents is critical for the real-world applicability of Deep RL, zero-shot policy transfer is still a challenging problem since even minor visual changes could make the trained agent completely fail in the new task. To address this issue, we propose a two-stage RL agent that first learns a latent unified state representation (LUSR) which is consistent across multiple domains in the first stage, and then do RL training in one source domain based on LUSR in the second stage. The cross-domain consistency of LUSR allows the policy acquired from the source domain to generalize to other target domains without extra training. We first demonstrate our approach in variants of CarRacing games with customized manipulations, and then verify it in CARLA, an autonomous driving simulator with more complex and realistic visual observations. Our results show that this approach can achieve state-of-the-art domain adaptation performance in related RL tasks and outperforms prior approaches based on latent-representation based RL and image-to-image translation.
\end{abstract}

\section{Introduction}
Deep reinforcement learning has been successful in a series of control problems, such as Atari 2600 video games \citep{mnih2013playing} and MuJoCo environments \citep{lillicrap2015continuous}. However, the advances of deep RL relies on a large amount of interactions with the environment. In addition, the policy tends to specialize to the training domain and fails to generalize to new domains even when these two domains are similar. It has been shown that slight visual changes on pixel-based observations from Atari games could cause the well trained policy totally break down \citep{gamrian2019transfer}. These two limitations make deep reinforcement learning algorithms inefficient when applied to sets of tasks. As a result, efficient domain adaptation approaches are important for the applicability of Deep RL.


Although state-of-the-art methods have demonstrated compelling performance in domain adaptation in RL, these approaches all have their limitations. Domain randomization \citep{tobin2017domain, andrychowicz2020learning, slaoui2020robust} relies on the availability of multiple source domains for training and cannot be applied in one-to-many generalization scenarios. Image-to-image translation approaches \citep{pan2017virtual, tzeng2020adapting, gamrian2019transfer} need a computationally expensive generator model for image translation. The extra burden on computation brought by the generator model is impractical for real-time applications such as autonomous driving. Other approaches utilize the latent embedding of encoder-decoder models to extract internal state representation for better generalization \cite{higgins2017darla}. However, domain-specific variations are also compressed into the latent embedding which could be problematic for zero-shot policy transfer.

To solve the problem of domain adaptation across related RL tasks and avoid limitations of prior methods, we propose to learn a latent unified state representation (LUSR) for different domains and then train RL agents in the source domain based on that. After the RL training, zero-shot policy transfer is evaluated in target domains. To learn LUSR, we split the latent state representation into domain-general embedding which contains information existing in all domains and domain-specific embedding that compress domain specific information. LUSR is composed of domain-general embedding only and thus is able to ignore domain-specific variations and generalize across domains. 

To empirically justify our approach, we conducted experiments in two car driving tasks with different visual complexity. We first applied our approach in CarRacing games with analysis of final domain adaptation performance, domain adaptation performance across the training period, generalization to totally unseen domains and policy explanation with saliency maps. Then we evaluated our approach in autonomous driving tasks in CARLA simulator \citep{Dosovitskiy17} with more challenging and realistic visual observations. 

In comparison with other approaches, LUSR does not need RL training in multiple source domains like domain randomization and thus is applicable to a wider range of tasks. In addition, LUSR does not need computationally expensive generator models and can achieve better training efficiency compared with image-to-image translation approaches that operate in pixel-space. Finally, in contrast with other approaches that use latent state representation, LUSR filters out the factors of variation across domains and ensures the latent state representation is unified across all domains. 
 



\section{Related Work}
Related work either tried to tackle domain adaptation in RL by directly generalizing the policy or learning generalized state representations.   

Domain randomization is the most popular approach to directly learn a policy with generalization capability \citep{tobin2017domain, andrychowicz2020learning, slaoui2020robust, laskin2020reinforcement}. By training on many source domains, the RL agent learns to ignore irrelevant factors of variation and attend to common features only. However, this approach relies on the availability of multiple source domains for training and the complexity of this approach scales with the number of variations. 

Instead of learning a policy with generalization capability directly, other works focus on the generalization of state representations. Some visual domain adaptation works use image-to-image translation to map the pixel-based states in the target domain to the paired states in the source domain \citep{pan2017virtual, tzeng2020adapting, gamrian2019transfer}. This is generally achieved via adversarial methods such as Generative Adversarial Networks (GANs) \citep{goodfellow2014generative}, and Unaligned GANs \citep{liu2017unsupervised, zhu2017unpaired} in the case where image pairs are lacking. While these methods provide promising results, the image translation brings extra burden during inference time which is impractical in real-time applications.  

Other works take one step further and try to learn a generalized state representation by mapping pixel-based states to a latent space \citep{higgins2017darla}. For example, the latent embedding of variational autoencoder (VAE) can be used as an internal latent state representation in RL. We call this method as VAE-Embedding. DARLA further extends the VAE to $\beta$-VAE to encourage the disentanglement of the latent embedding and uses one internal layer of a pre-trained Denoising AutoEncoder (DAE) \citep{vincent2010stacked} as the reconstruction target. Although disentanglement in latent state representation makes it easier for RL agents to ignore irrelevant domain-specific features, the policy transfer performance is not guaranteed because domain-specific features still reside in the latent state representation and their contribution to the policy output cannot be generalized to other domains. CURL extracts high-level features from raw pixels using contrastive learning and greatly improves the sample efficiency \citep{laskin2020curl}.

\begin{figure*}[ht]
\begin{subfigure}{0.48\textwidth}
\centering
\includegraphics[width=0.8\linewidth]{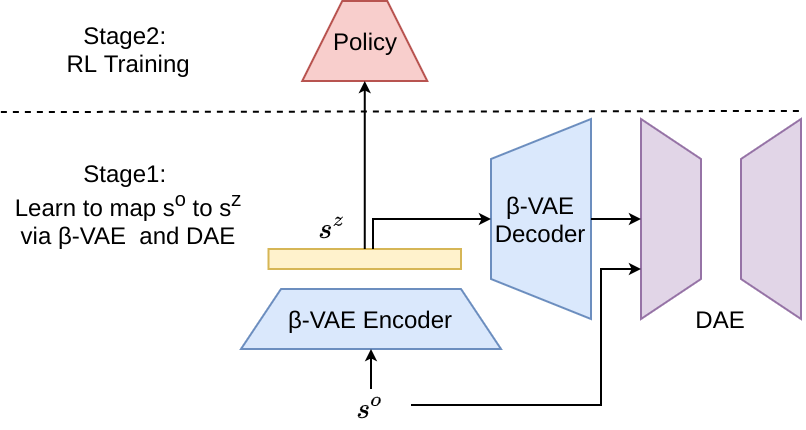}
\caption{DARLA}
\end{subfigure}
\hfill
\begin{subfigure}{0.5\textwidth}
\centering
\includegraphics[width=0.8\linewidth]{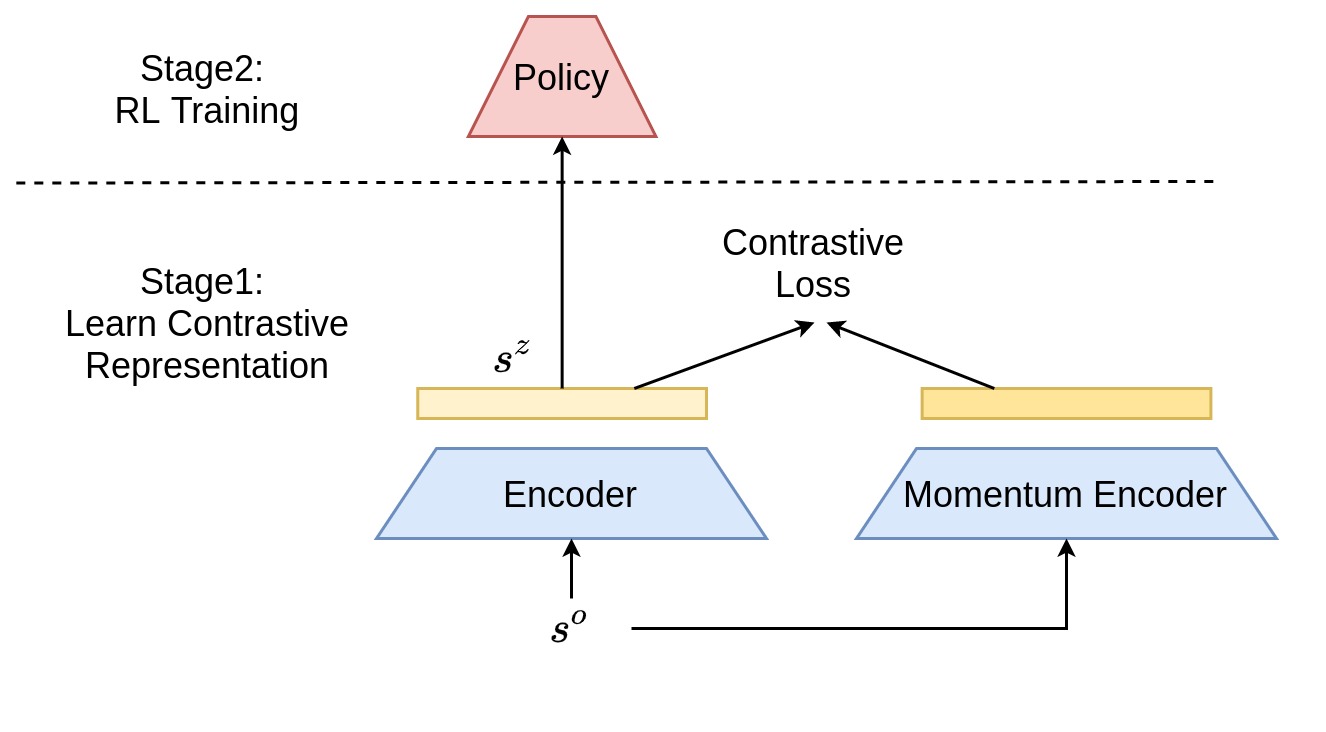}
\caption{CURL}
\end{subfigure}
\newline
\begin{subfigure}{0.48\textwidth}
\centering
\includegraphics[width=0.8\linewidth]{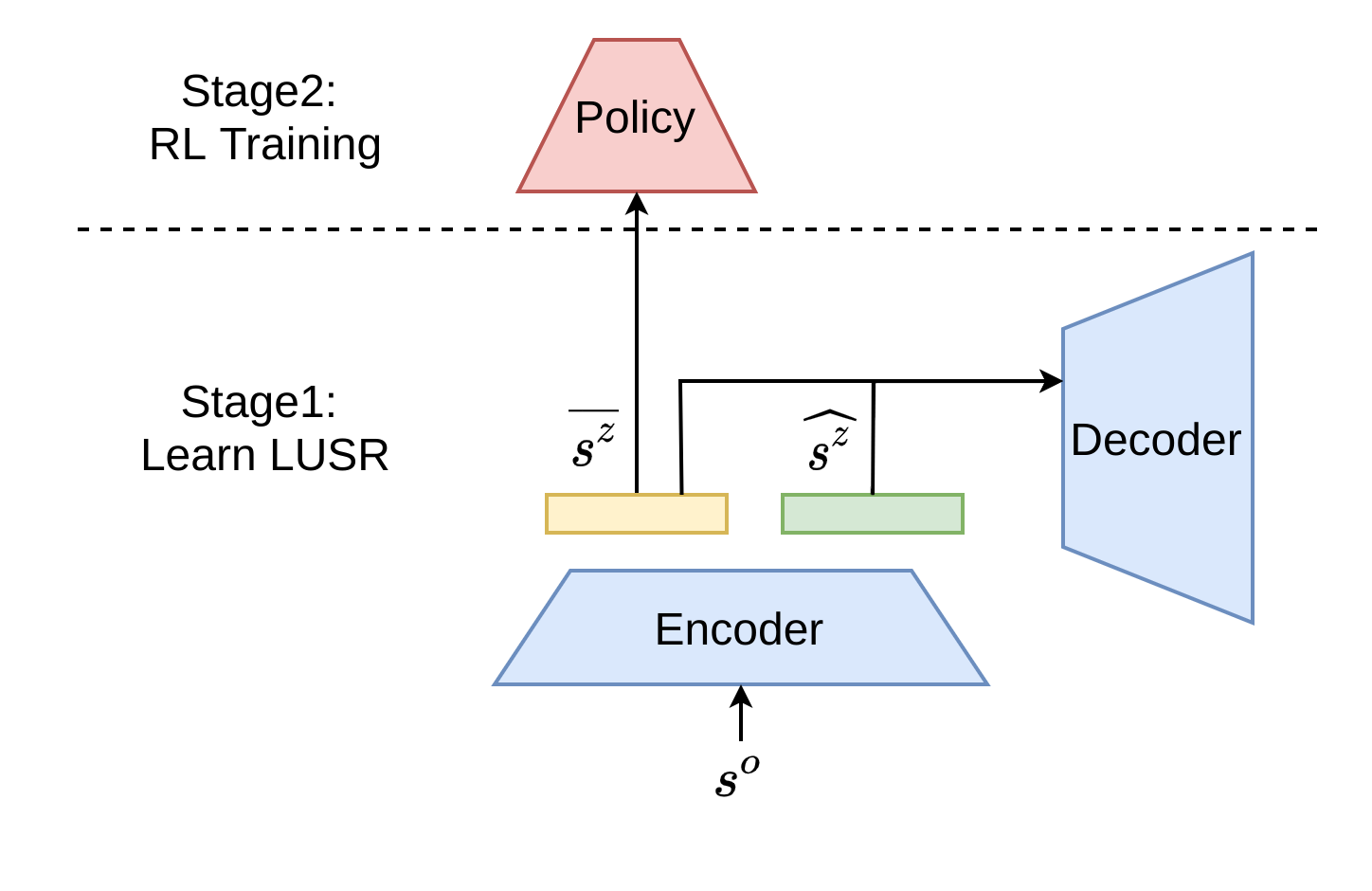}
\caption{LUSR (ours)}
\end{subfigure}
\hfill
\begin{subfigure}{0.49\textwidth}
\centering
\includegraphics[width=0.8\linewidth]{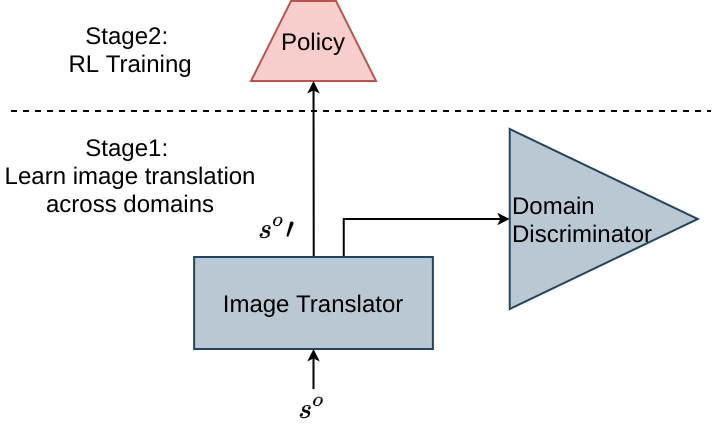}
\caption{CycleGAN Image Translation}
\end{subfigure}
\caption{Architectures of our method (LUSR) and other benchmarks (DARLA, CURL and CycleGAN based image-to-image translation) used in this work for comparison. The architecture of VAE-Embedding could be considered as a special case of DARLA that replaces $\beta$-VAE with VAE and avoids the usage of DAE. The learning of all these approaches could be divided into two stages. The first stage is learning appropriate state representations that support domain adaptation in RL and the second stage is doing RL training.}
\label{fig:method}
\end{figure*}

In this work, we choose VAE-Embedding, DARLA, CURL and CycleGAN-based image-to-image translation as benchmarks. To make it more clear how LUSR differs from them, we use Figure \ref{fig:method} to demonstrate their frameworks.

\section{Domain Adaptation In Reinforcement Learning}

Reinforcement learning is an area that studies how agents should take actions in an environment in order to maximize their cumulative rewards. The environment is typically stated in the form of a Markov decision process (MDP), which is expressed in terms of the tuple ($S, A, T, R$) where $S$ is the state space, $A$ is the action space, $T$ is the transition function and $R$ is the reward function. At each time step $t$ in the MDP, the agent takes an action $a_t$ in the environment based on current state $s_t$ and receives a reward $r_{t+1}$ and next state $s_{t+1}$. The goal of the agent is to find a policy $\pi(s)$ to choose actions that maximize the discounted cumulative future rewards $r_t + \gamma r_{t+1} + \gamma^2 r_{t+2} + ...$, where $\gamma$ is the discount factor ranging from $0$ to $1$.   

To formalize domain adaptation scenarios in the setting of reinforcement learning, we define the source and target domains as $D_S$ and $D_T$. Each domain corresponds to a MDP defined as tuple ($S, A, T, R$) and thus the MDPs in the source domain $D_S$ and target domain $D_T$ are defined as ($S_S, A_S, T_S, R_S$) and ($S_T, A_T, T_T, R_T$), respectively. The source and target domains could have distinct state spaces $S$, but their action spaces $A$ should be the same and their transition function $T$ and reward function $R$ should have similarity because of the sharing internal dynamics. Namely, we focus on policy transfers where $T_S \approx T_T$, $R_S \approx R_T$, $A_S = A_T$, but $S_S \neq S_T$. 

Take autonomous driving as an example, different domains may correspond to different weather conditions. For instance, the source domain is driving on a sunny day and the target domain is driving on a rainy day. While state space $S$ (visual observations) could differ due to rain and different lighting conditions, the action space $A$ (throttle and steering) remains the same. As for the transition function $T$ and reward function $R$, they should have similarity since the state transition for both domains are governed by the traffic condition and driving control while the reward function for both domains are determined by the movement of the vehicle.

\section{Methods}
Our approach focuses on learning a latent unified state representation (LUSR) for states from different domains in RL. In this section, we first introduce the definition of LUSR and then introduce how to learn it.



\subsection{LUSR Definition}

We first introduce two notions for state space in RL which are the agent's raw observation state space $S^o$ and the agent's internal latent state space $S^z$. Raw observation states $s^o$ consists of a grid of pixels while each unit in the internal latent state $s^z$ represents a high level semantic feature. A mapping function $\mathcal{F}: S^o \to S^z$ maps the observation state to the corresponding internal latent state. 
In our work, high level semantic features in $S^z$ are further divided into domain-specific ones (such as weather conditions in the driving task) and domain-general ones (such as vehicle dynamics). Here we denote $S^z = (\widehat{S^z}, \overline{S^z})$ where $\widehat{S^z}$ represents domain-specific features and $\overline{S^z}$ represents domain-general features. For state representation in source and target domains, this is summarized as

\begin{equation}
\begin{aligned}
      S^o_S  & \neq S^o_T \\
    S^z_S = (\widehat{S^z_S}, \overline{S^z_S}); & \quad S^z_T = (\widehat{S^z_T}, \overline{S^z_T}) \\
    \overline{S^z_S} =  \overline{S^z_T}; & \quad \widehat{S^z_S} \neq \widehat{S^z_T}  
\end{aligned}
\end{equation}

In our setting of domain adaptation, the transition function $T$ and reward function $R$ only depend on $\overline{S^z}$ which is consistent across domains. Here we define the reward and transition function that take $s^o$ as input as $R^o$ and $T^o$ while the reward and transition function that take $s^z$ as input as $R^z$ and $T^z$. Then, we have

\begin{equation}
\begin{aligned}
      {T^o_S} \neq {T^o_T}; &\quad {R^o_S} \neq {R^o_T} \\
      {T^z_S} = T(\overline{S^z_S}) & = T(\overline{S^z_T}) = {T^z_T} \\
      {R^z_S} = R(\overline{S^z_S}) & = R(\overline{S^z_T}) = {R^z_T}  \\
\end{aligned}
\end{equation}


Since $\overline{S^z}$ is consistent across domains and the reward structure ($T$ and $R$) depend only on this representation (not on $\widehat{S^z}$), the RL agent taking $\overline{S^z}$ as input will be able to be trained successfully and the trained agent also has the capability to adapt from the source domain to target domains. As a result, the goal of our approach is learning the mapping function $\mathcal{F}: S^o \to \overline{S^z}$ that maps raw observation states to the latent unified state representation which we call LUSR.

\subsection{Learning LUSR}
In this work, we choose to learn the mapping function $\mathcal{F}: S^o \to \overline{S^z}$ via Cycle-Consistent VAE \citep{jha2018disentangling} which is a non-adversarial approach to disentangle domain-general and domain-specific factors of variation. Similar to VAE \citep{kingma2013auto}, Cycle-Consistent VAE is also composed of an encoder and a decoder. However, the output from the encoder is split into domain-general and domain-specific embeddings. To learn the mapping function $\mathcal{F}$, a number of random observation states from a set of pre-defined domains are first collected and then used as input for Cycle-Consistent VAE model training. Once the model is trained, the encoder is able to map observation states $s^o$ from any domain in the domain set to a latent state representation composed of $\overline{s^z}$ and $\widehat{s^z}$. As a result, we use the trained encoder as our mapping function $\mathcal{F}$ and keep only domain-general representation as LUSR.  


Cycle-Consistent VAE is based on the idea of cycle consistency whose intuition is that two well trained forward and reverse transformations composed together \emph{in any order} should approximate an identity function. For example, in the VAE, the encoder is a forward transformation that converts an input image to a latent vector while the decoder is the reverse transformation that converts the latent vector back to a reconstructed image. Here we define the forward cycle as: $Dec(Enc(s^o)) = s^{o}\prime $ and the reverse cycle as $Enc(Dec(\widehat{s^z}, \overline{s^z})) =(\widehat{s^z \prime}, \overline{s^z \prime}) $. As indicated by the cycle consistency, $s^{o}\prime$ should be close to $s^{o}$ and also $(\widehat{s^z \prime}, \overline{s^z \prime})$ should be close to $(\widehat{s^z}, \overline{s^z})$. 

In the forward cycle of Cycle-Consistent VAE, for two observation states $s^o_1$, $s^o_2$ from the same domain, $Enc(s^o_1) = \widehat{s^z_1}, \overline{s^z_1}$ and $Enc(s^o_2) = \widehat{s^z_2}, \overline{s^z_2}$. Since both originate from the same domain and $\widehat{s^z}$ contains only domain-specific information, swapping $\widehat{s^z_1}$ and $\widehat{s^z_2}$ should have no effect on the reconstruction loss which means we should get $Dec(\widehat{s^z_2}, \overline{s^z_1}) \approx s^o_1$ and $Dec(\widehat{s^z_1}, \overline{s^z_2}) \approx s^o_2$. This operation ensures that domain-specific information and domain-general information are compressed into $\widehat{s^z}$ and $\overline{s^z}$ separately.

In the reverse cycle, a randomly sampled $\overline{s^z}$ is passed through the decoder in combination with two domain-specific embeddings $\widehat{s^z_1}$ and $\widehat{s^z_2}$ to obtain two reconstructed images $s^o_1\prime$ and $s^o_2\prime$. Since both $s^o_1\prime$ and $s^o_2\prime$ are generated based on the same $\overline{s^z}$, their corresponding domain-general latent embedding $\overline{s^z_1}\prime$ and $\overline{s^z_2}\prime$ should also be the same.

As a result, the objective for Cycle-Consistent VAE to minimize is 
\begin{equation}
\mathcal{L}_{cyclic} = \mathcal{L}_{forward} + \mathcal{L}_{reverse}
\end{equation}
where 
\begin{equation*}
\begin{aligned}
\mathcal{L}_{forward} = &  -\mathbb{E}_{q_\phi(\overline{s^z}, \widehat{s^z} \mid s^o)} [\log p_\theta(s^o \vert \overline{s^z}, \widehat{s^z\ast})] \\  &  + KL(q_\phi(\overline{s^z} \vert s^o) || p(\overline{s^z})) \\
\mathcal{L}_{reverse}  = &  \mathbb{E}_{\overline{s^z} \sim p(\overline{s^z})} [|| \overline{ q_{\phi}} (p_{\theta}(\overline{s^z}, \widehat{s^z_1})) - \overline{q_{\phi}} (p_{\theta}(\overline{s^z}, \widehat{s^z_2}))  ||_1]
\end{aligned}
\end{equation*}

 $\mathcal{L}_{forward}$ here is a modified variational
upper-bound and $\mathcal{L}_{reverse}$ is the loss for cycle consistency. $q_\phi$ and $p_\theta$ are parameterized functions of the encoder and decoder. We define $\overline{q_\phi}$ as $q_\phi$ that only keeps the domain general embedding as output. The latent embedding $s^z$ is composed of 
 $\overline{s^z}$ and $\widehat{s^z}$ which are domain-general and domain-specific latent embeddings corresponding to observation state $s^o$. $\widehat{s^z\ast}$ represents any random domain-specific embedding from the same domain while $\widehat{s^z_1}$ and $\widehat{s^z_2}$ are two different domain-specific embeddings.

\section{Experiments}
We first apply our approach to a set of CarRacing variants which allows manual manipulations of the visual observations. This flexibility allows us to analyze the influence of different categories of variations on the performance of domain adaptation in RL. After that, our approach is applied in autonomous driving tasks in the CARLA simulator in which the observational states are much more complicated and helps us to evaluate the ability of our approach to scale up to more challenging tasks. 

\subsection{CarRacing}

\begin{figure}
    \centering
    \includegraphics[width=0.95\linewidth]{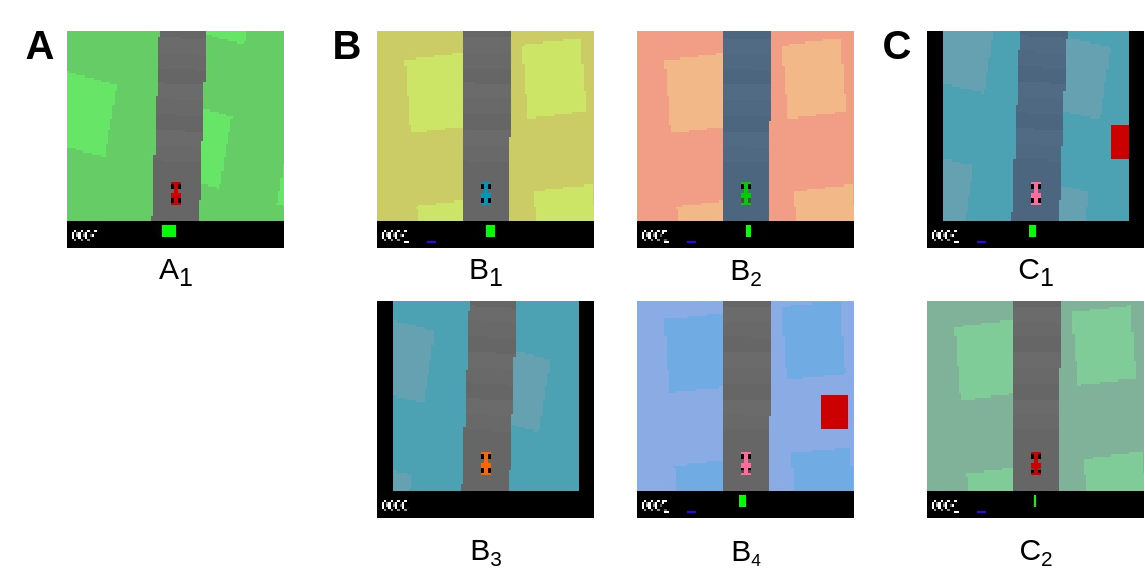}
    \caption{Variants of CarRacing games. \textbf{A}. The original version of CarRacing game which is set as the source domain. \textbf{B}. The seen target domains of CarRacing games whose observation states are collected for learning LUSR. \textbf{C}. The unseen target domains of CarRacing games. These two domains are never exposed to the agent, not only during RL training but also during latent state representation learning.}
    \label{fig:carracing}
\end{figure}

 We first apply our approach on variants of CarRacing game which is a continuous control task to learn to drive from pixels. As shown in Figure \ref{fig:carracing}, we divide all variants into three categories: source domain, seen target domains and unseen target domains. We first collect random observation states from the source domain and seen target domains to learn the mapping function $\mathcal{F}$ which maps raw observation states to LUSR. In each domain, we collect 100k images and thus have 500k images in total (one source domain and four seen domains). The collected images are used as the dataset to train a Cycle-Consistent VAE model whose encoder is the mapping function $\mathcal{F}$ we need. After that, we train the RL agent in the source domain with LUSR for 10 millions steps via Proximal Policy Optimization (PPO) \citep{schulman2017proximal} algorithm. In this work, we use Ray RLlib \citep{liang2018rllib} and RLCodebase \citep{rlcodebase} for the PPO implementation. After the RL training, we test the RL agent's performance of adapting to the seen target domains and unseen target domains. 

With the ability of inducing manual manipulations over the observation states, we design two types of variations. The first type is color change including changing the background color, the car color and the road color. For example, the background color in all target domains is different from that in the source domain. Another type of variation is inducing patterns. For example, we induce a red blob at a fixed position in the fourth game of seen target domains ($B_4$). In summary, compared with the source domain $A_1$, seen target domains $B_1$ and $B_2$ have color changes while $B_3$ and $B_4$ have both color changes and new patterns. For unseen target domains, $C_1$ combines all variations introduced in seen target domains and $C_2$ uses a totally new background color.

\subsection{Autonomous Driving In CARLA}
\begin{figure}
    \centering
    \includegraphics[width=1\linewidth]{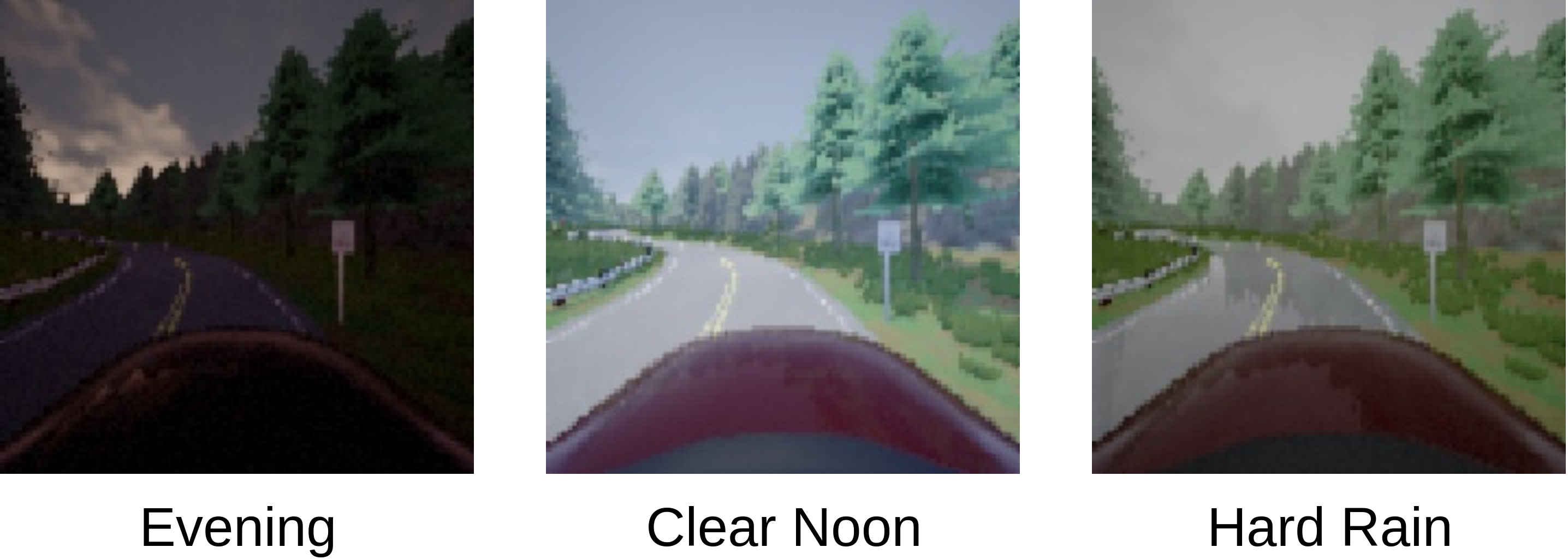}
    \caption{Experiment of the driving task in CARLA simulator. Examples of the driver view (observation states) under three different weather conditions: evening, clear noon and hard rain from left to right. }
    \label{fig:carla}
\end{figure}

Although CarRacing games are suitable to study the domain adaptation problem of RL agents, the observation states are relatively simple compared to real world observations during driving. To further evaluate the performance of our approach, we applied it in a much more challenging task: autonomous driving in the CARLA simulator. In this experiment, we first choose a start point and an end point in the map of town07 for the driving task. To go from the start point to the destination, the vehicle must go through a curvy road and avoid collisions and lane crossings. The action space is composed of two continuous values for driving control (throttle and steering). At each step, the driving control is applied on the vehicle for 0.1 simulation second. We use images captured by a camera attached to the front end of the RL agent vehicle along with the current speed as the observation states. Each episode terminates if the vehicle collides, runs out of the lane, reaches the destination, or reaches the maximum episode timesteps (800 in this experiment). To make the CARLA simulator compatible with RL training, we use a gym wrapper of CARLA in the experiment \citep{chen2019model}. 

To study domain adaptation, we test the model under different weather conditions and times of day. Specifically, the RL agent is first trained in the late evening and then tested in the weather of clear noon and hard rain. Examples of the driver view under these conditions are shown in Figure \ref{fig:carla}. 

Since CARLA aims to provide realistic simulations of urban driving, the observation states in this driving task are much more complex and challenging for domain adaptation compared to states in CarRacing games. Besides that, the complexity of environment dynamics also makes the simulation of CARLA slower compared to CarRacing games. As a result, we set the number of PPO training steps in this experiment as 50k. This further requires the RL agent to have a high training efficiency to achieve good performance with limited number of interactions with the environment. 

\section{Results and Discussion}
In this section, we introduce the results of our approach in two experiments along with other benchmarks. 
\subsection{CarRacing}

\subsubsection{LUSR Demonstration}
\begin{figure}
    \centering
    \includegraphics[width=1\linewidth]{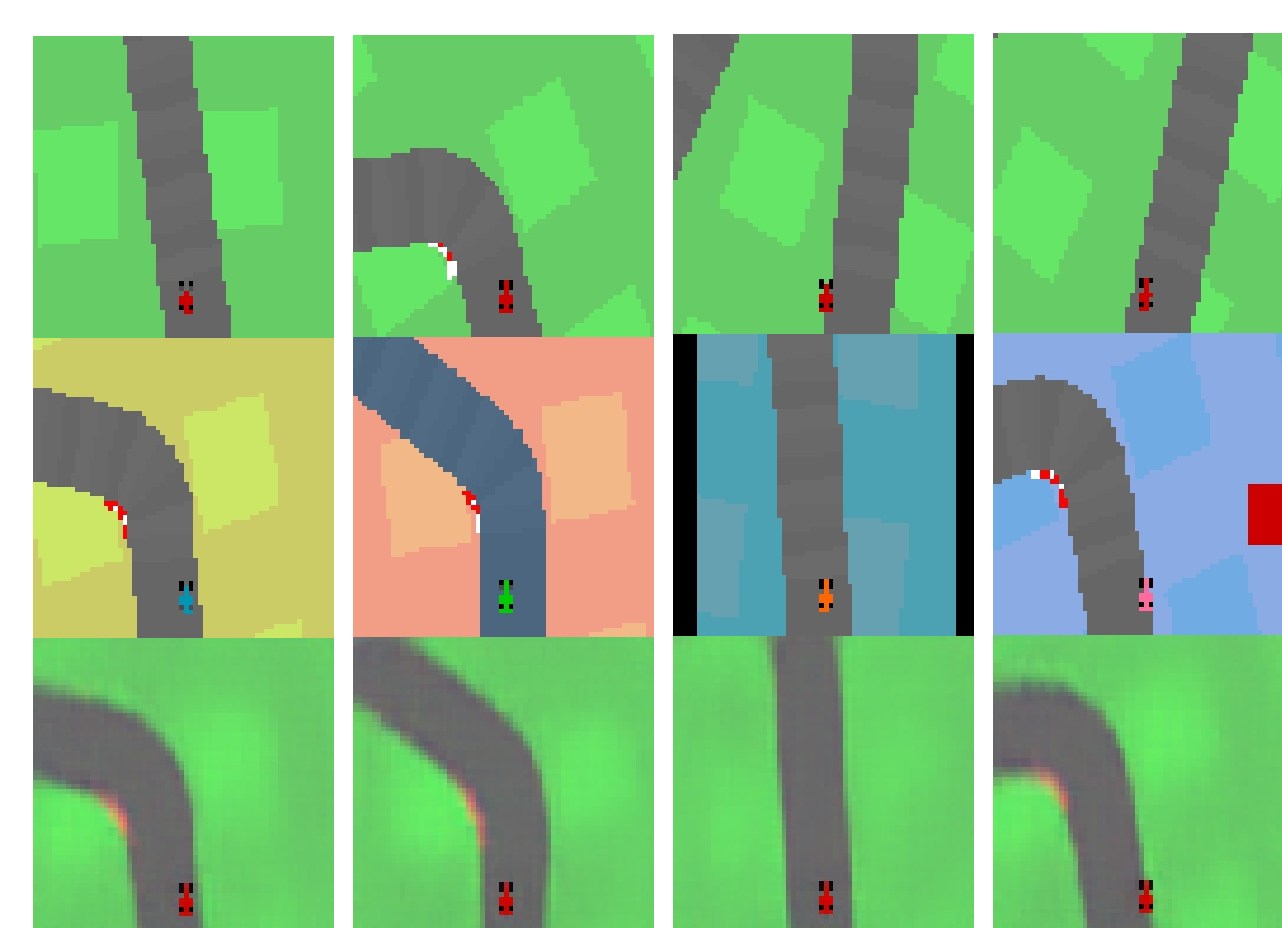}
    \caption{Results of Cycle-Consistent VAE. The first row are four random images from the source domain and the second row are four random images from four seen target domains respectively. The last row are reconstructed images that take $\widehat{s^z}$ from the first row and $\overline{s^z}$ from the second row.} 
    \label{fig:cyclevae_recon}
\end{figure}

We first demonstrate the effectiveness of LUSR. In our approach, the latent embedding is split into domain-general embedding $\overline{s^z}$ and domain-specific embedding $\widehat{s^z}$. 
To verify that these two embeddings are well disentangled, we first select random images from the source domain and seen target domains and then extract their latent embeddings. 
For example, we get $\widehat{s^z_1}$ and $\overline{s^z_1}$ for image $s^o_1$, and $\widehat{s^z_2}$ and $\overline{s^z_2}$ for image $s^o_2$.
If we feed the decoder with a latent embedding composed of $\widehat{s^z_1}$ and $\overline{s^z_2}$, the reconstructed image $s^o\prime$ should have visual features from both $s^o_1$ and $^o_2$. 
Furthermore, the shared features between $s^o\prime$ and $s^o_1$ should be domain-specific while the shared features between $s^o\prime$ and $s^o_2$ should be domain-general. As shown in Figure \ref{fig:cyclevae_recon}, the third row of images are generated with $\widehat{s^z}$ from the first row and $\overline{s^z}$ from the second row. As a result, their domain-specific features (color and patterns) are the same as the first row of images while the domain-general features (road shape) are the same as images in the second row. 


\subsubsection{Domain Adaptation After Training}
After the RL training in the source domain, we evaluate the domain adaptation performance of our approach and other benchmarks in both seen target domains and unseen target domains (see table \ref{tab: carracing}). The result shows that RL agents trained with LUSR are able to generalize to all target domains almost without performance loss and achieve best scores in most target domains. For DARLA, the choice of parameter $\beta$ strongly affects the adaptation performance. Furthermore, it generalizes better in target domains with only color variations and could fail to adapt to domains with new patterns. VAE-Embedding has notable performance loss for all target domains. CURL has the worst transfer performance among all approaches and completely fails in most domains. Finally, CycleGAN can also adapt to all target domains without performance loss. However, the final scores are not comparable to other approaches that use latent embeddings as input for RL training.  

\begin{table*}[]
\scalebox{0.85}{
\begin{tabular}{c|c|cccc|cc}
\hline
Approach      & Source Domain & \multicolumn{4}{c}{Seen Target Domains}                       & \multicolumn{2}{c}{Unseen Target Domains} \\ 
                \hline 
           & \makecell{CarRacing\_A1 \\ Score}  & \makecell{CarRacing\_B1 \\ Score(Ratio)} & \makecell{CarRacing\_B2 \\ Score(Ratio)} & \makecell{CarRacing\_B3 \\ Score(Ratio)} & \makecell{CarRacing\_B4 \\ Score(Ratio)} & \makecell{CarRacing\_C1 \\ Score(Ratio)}  & \makecell{CarRfacing\_C2 \\ Score(Ratio)}\\ \cline{2-8}
LUSR & 805.13 & 803.52 (\textbf{1.00}) & \textbf{807.37} (\textbf{1.00}) & \textbf{803.11} (\textbf{1.00}) & \textbf{781.94} (\textbf{0.97})  & \textbf{678.7} (\textbf{0.84}) & \textbf{800.56} (\textbf{0.99}) \\
DARLA($\beta=10$) & 845.87 & 645.81 (0.76) & 250.85 (0.30) & -72.99 (-0.09) & -62.96 (-0.07) & -65.72 (-0.08)  & 631.09 (0.75)\\
DARLA($\beta=30$) &851.48 & \textbf{834.99} (\textbf{0.98}) & \textbf{819.05} (\textbf{0.96}) & -60.76 (-0.07) & -73.18 (-0.09)  & -72.55 (-0.09) & \textbf{806.76} (\textbf{0.95})\\
DARLA($\beta=100$) & 778.78 & 704.27 (0.90) & 207.9 (0.27) & 451.81 (0.58) & 27.77 (0.04)  & 182.35 (0.23) & 539.63 (0.69)\\
VAE-Embedding & 816.74 & 616.89 (0.76) & 282.71 (0.35) & 484.57 (0.59) & 223.88 (0.27) & 332.58 (0.41)  & 595.42 (0.73)\\
CURL & 748.58 & 560.23(0.75) & -44.24(-0.06) & -55.29(-0.07) & -32.45(-0.04) & -113.13(-0.15) & -69.23(-0.09) \\
CycleGAN &709.12 & 707.64 (\textbf{1.00}) & 704.33 (\textbf{0.99}) & 713.86 (\textbf{1.01}) & 711.85 (\textbf{1.00}) & \textbf{715.43}(\textbf{1.01}) & 671.67(\textbf{0.96}) \\
\end{tabular}}
\caption{Domain adaptation performance of LUSR and benchmarks in CarRacing games. We train 3 models for each approach and evaluate each model for 100 episodes in each domain after training. The average final score of 3 models are reported in the table for each approach. We also report the ratio of scores achieved in target domains to the score achieved in the source domain to demonstrate the policy transfer performance.}
\label{tab: carracing}
\end{table*}

\begin{table*}[]
\centering
\scalebox{0.85}{
\begin{tabular}{c|cc|cccc}
\hline
Approach & \multicolumn{2}{c|}{Source Domain} & \multicolumn{4}{c}{Target Domains} \\ \hline
 & \multicolumn{2}{c|}{CARLA (Evening)} & \multicolumn{2}{c|}{CARLA (Clear Noon)} & \multicolumn{2}{c}{CARLA (Hard Rain)} \\ \cline{2-7} 
 & Score & Steps & Score & \multicolumn{1}{c|}{Steps} & Score & Steps \\
LUSR & \textbf{1125.06} & 469.1 & \textbf{1175.61} & \multicolumn{1}{c|}{565.3} & \textbf{1270.32} & 515.6 \\   
DARLA & 841.59  & 342.0 & 194.41 & \multicolumn{1}{c|}{134.2} & 187.97 & 119.9 \\
VAE-Embedding & 1113.90 & 384.9 & 674.42 & \multicolumn{1}{c|}{744.2} & 846.14 & 527.4 \\
CURL &  1112.42 & 521.2 &  44.34 &  \multicolumn{1}{c|}{42.4} & 73.63 & 60.2 \\
CycleGAN & 333.57 & 175.1 & 333.88 & \multicolumn{1}{c|}{174.9} & 332.71 & 163.6
\end{tabular}}
\caption{Domain adaptation performance of LUSR and benchmarks in CARLA autonomous driving tasks. We train 3 models for each approach and choose the best model for evaluation. Each model is evaluated for 10 episodes. The average score and time steps spent in each episode are reported in the table.}
\label{tab: carla_result}
\end{table*}

\subsubsection{Domain Adaptation During Training}
\begin{figure}
    \centering
    \includegraphics[width=0.8\linewidth]{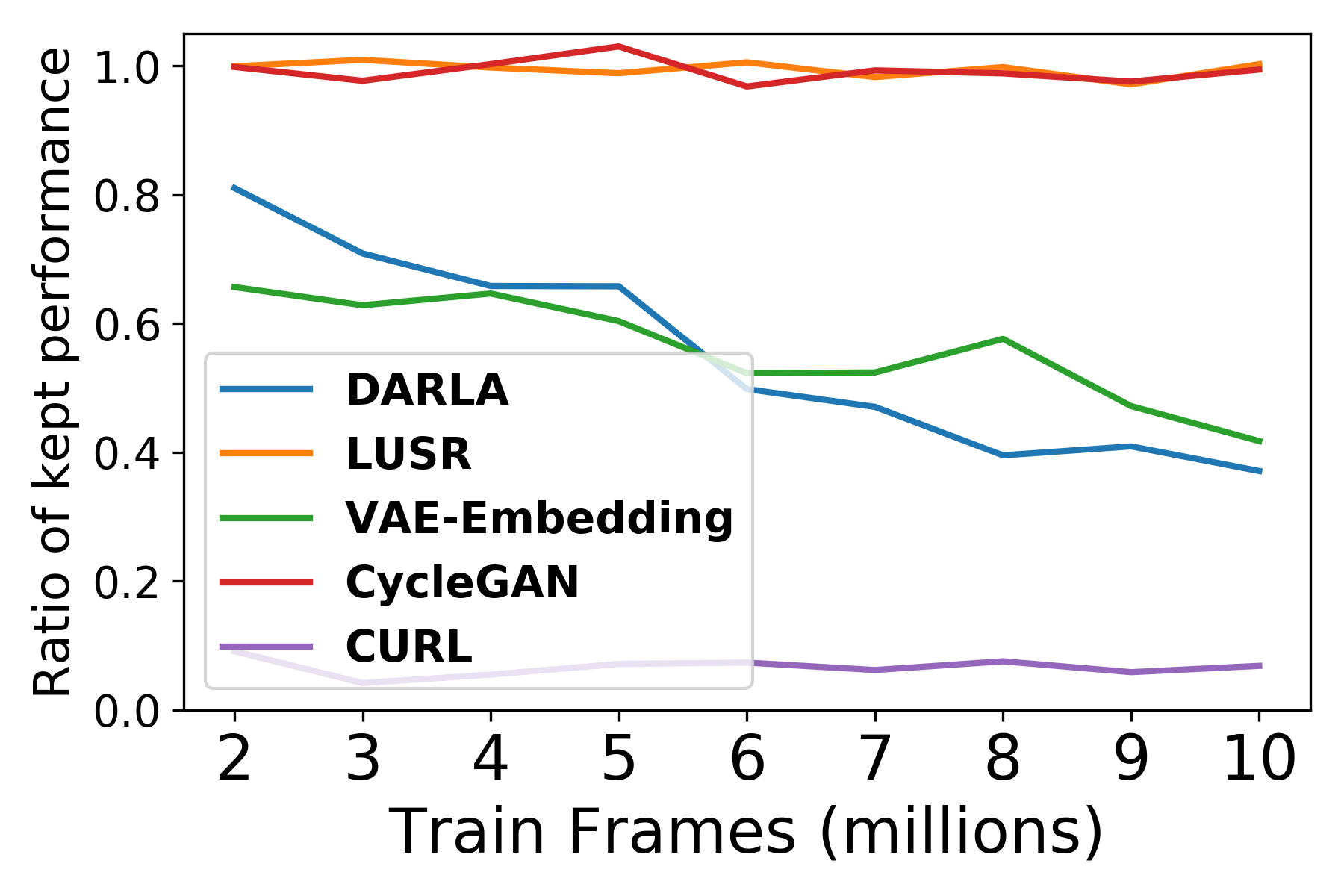}
    \caption{Domain adaptation performance during training in CarRacing comparing LUSR to other benchmarks.}
    \label{fig:duringtrain}
\end{figure}

Besides the domain adaptation performance after training, we're also interested in the adaptation performance during the training period. Since the RL agent will be more and more deterministic in action selection during the training, the domain adaptation performance could also be affected. As a result, we evaluate the model adaptation performance every 1 million frames of training for all approaches. Our result shows that both LUSR and CycleGAN have consistent adaptation performance during the whole training period while the adaptation performance of DARLA and VAE-Embedding gradually decrease (see Figure \ref{fig:duringtrain}). This demonstrates that domain-specific features do contribute to the RL policy output if they are included in the latent state representation and their influence will be more and more problematic as RL training goes on.


\subsubsection{Saliency Map}
\begin{figure}
    \centering
    \includegraphics[width=1\linewidth]{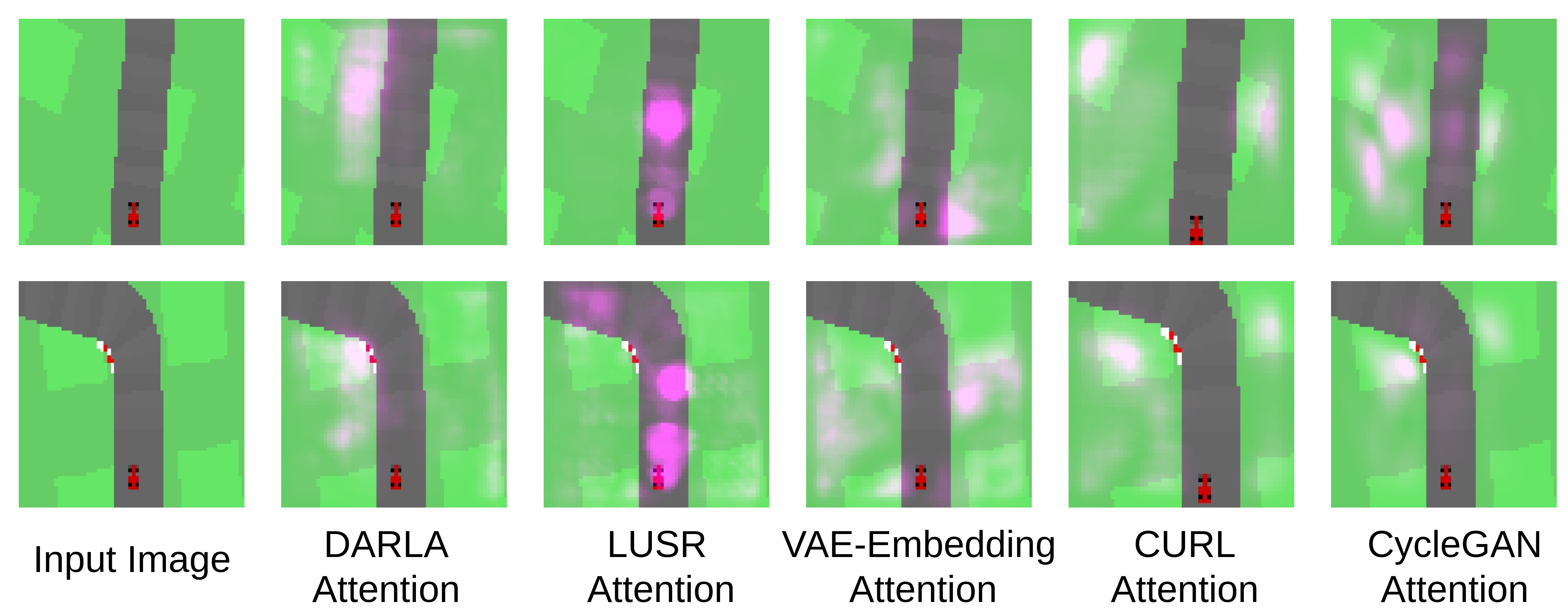}
    \caption{Examples of saliency maps generated by RL agents trained via DARLA, LUSR, VAE-Embedding, CURL and CycleGAN. The RL agent trained with LUSR has the most centralized attention and mainly attends to the center of the road.}
    \label{fig:saliency}
\end{figure}

Saliency map is an approach to visualize and understand the behavior of RL agents. In this work, we also use saliency maps \citep{greydanus2018visualizing} to visualize how RL agents trained with different methods attend to the observation states (see Figure \ref{fig:saliency}). The result shows that RL agent trained with LUSR has more centralized attention and mainly attends to the center of the road. In comparison, the saliency maps generated by other approaches are much more diffused and attend more to the edges between road and grass rather than road itself. Although it also makes sense to learn to drive based on the edges, the contrast between them could also change when the color of grass changes and thus brings more challenges in generalization. This may explain why LUSR has better generalization performance than other benchmarks.   

\begin{figure*}[h]
\begin{tabular}[t]{cc}
    \begin{subfigure}{0.68\textwidth}
    \centering
    \includegraphics[width=1\linewidth]{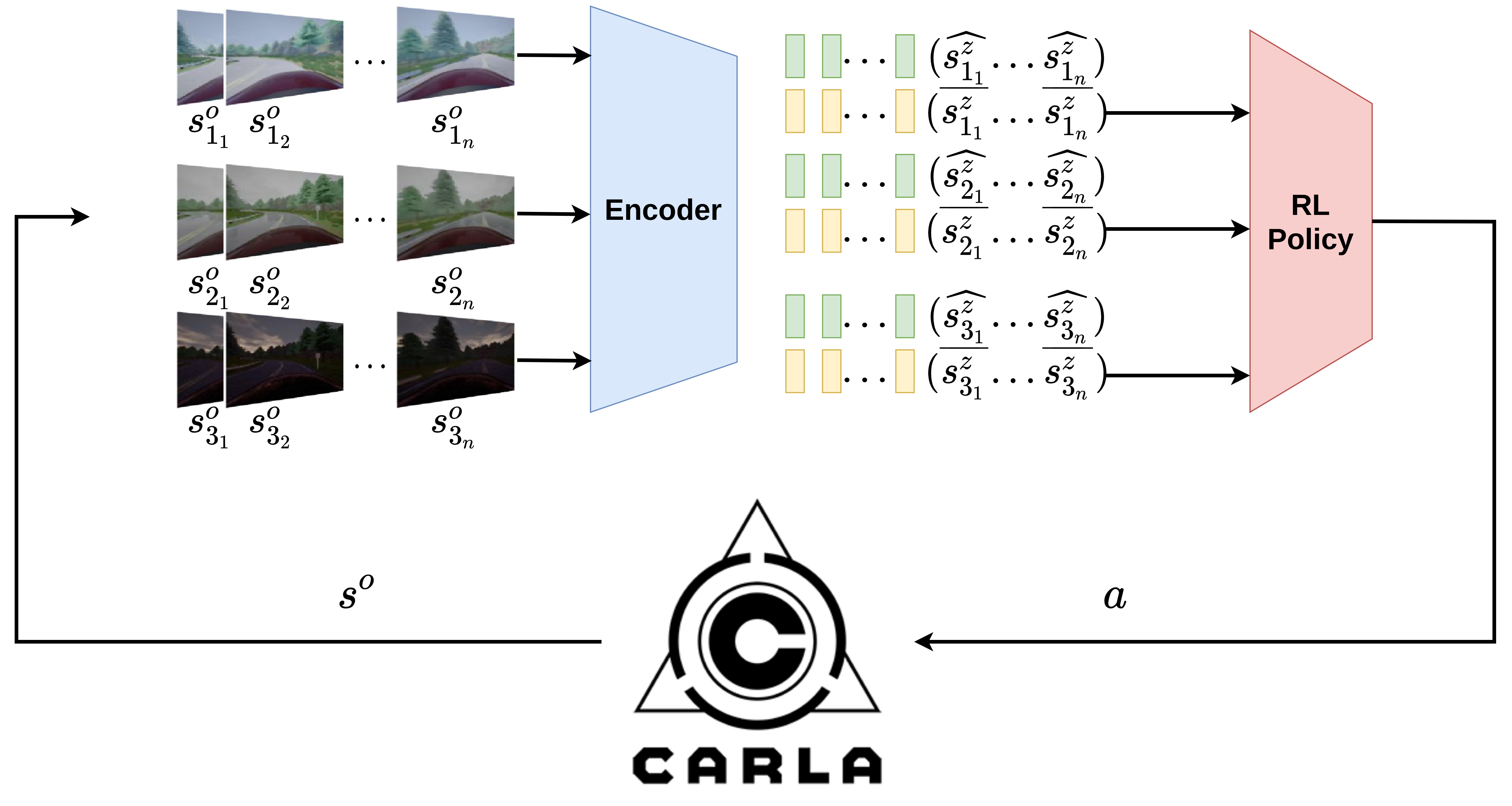}
    \caption{}
    \end{subfigure}
    &
            \begin{tabular}{c}
        \smallskip
            \begin{subfigure}[t]{0.3\textwidth}
                \centering
                \includegraphics[width=0.8\textwidth]{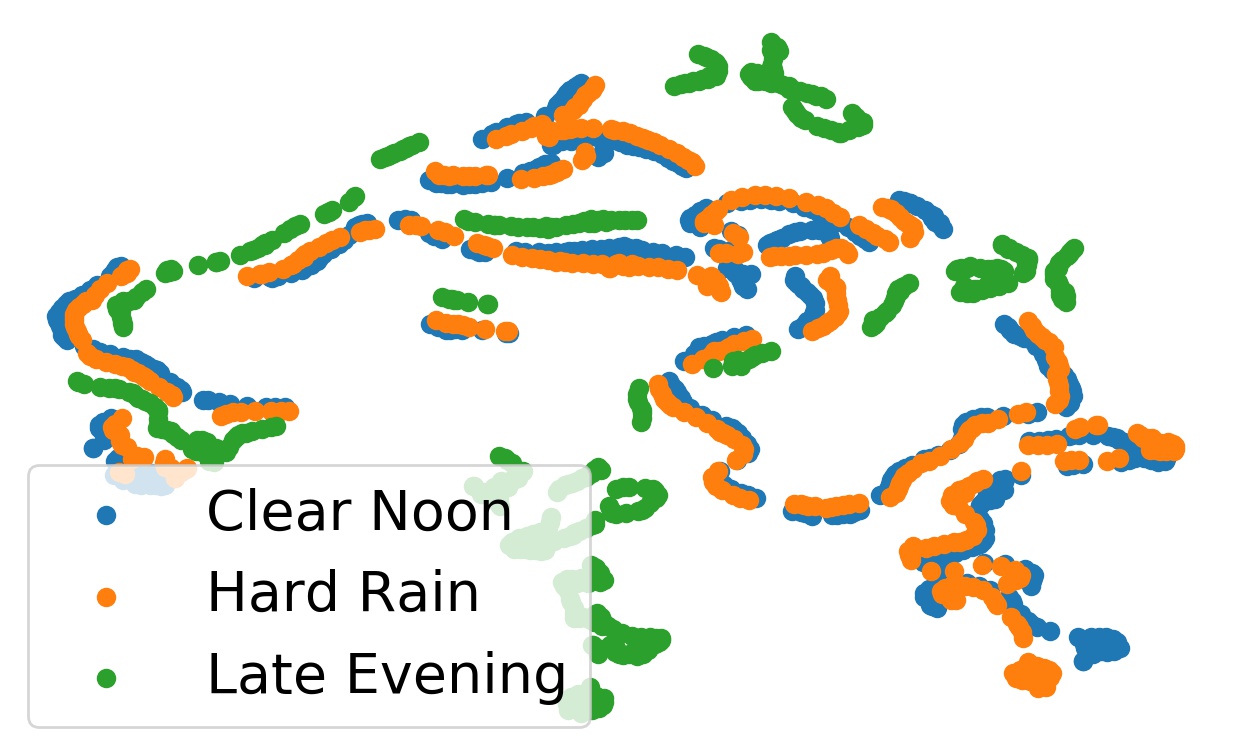}
                \caption{}
                \label{fig:tsne1}
            \end{subfigure}\\
            \begin{subfigure}[t]{0.3\textwidth}
                \centering
                \includegraphics[width=0.8\textwidth]{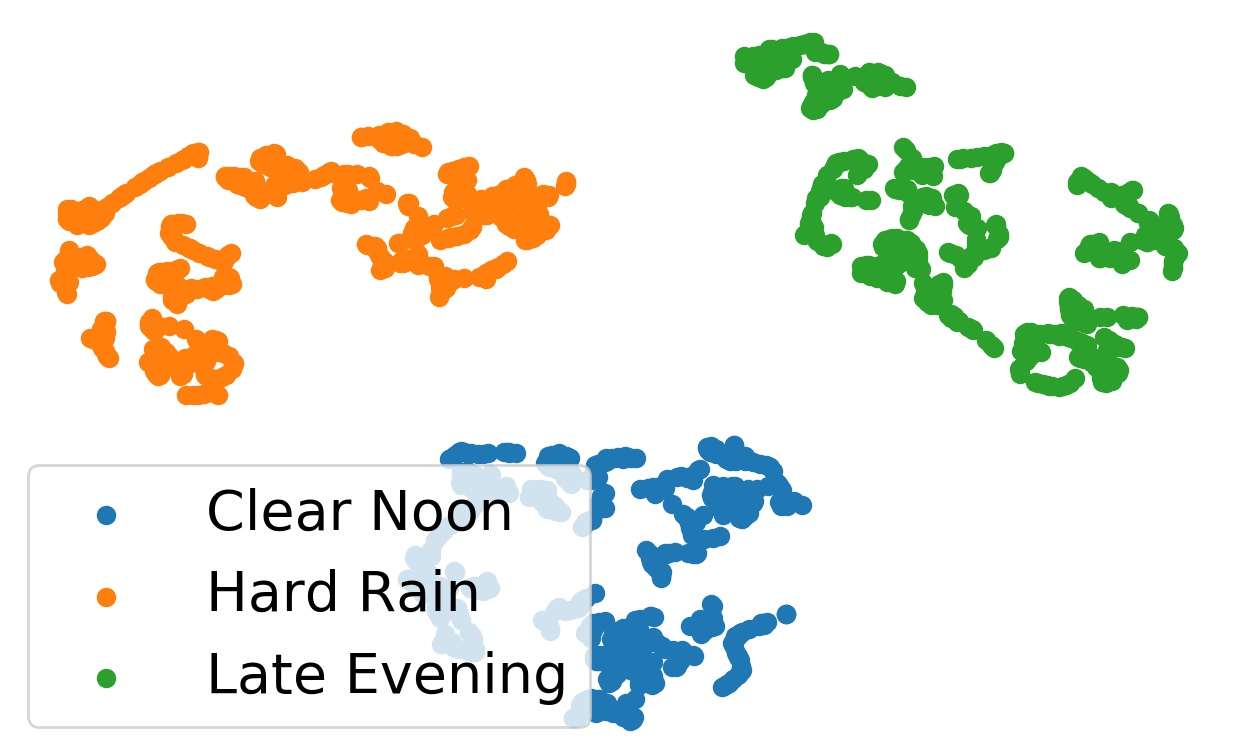}
                \caption{}
                \label{fig:tsne2}
            \end{subfigure}
        \end{tabular}\\
\end{tabular}
\caption{Demonstration of the disentanglement of domain-general embedding and domain-specific embedding in related CARLA driving tasks. \textbf{a}. The workflow of generating paired observational states and extracting latent embeddings. \textbf{b},\textbf{c}. t-SNE plot of the domain-general and domain-specific embeddings from three CARLA driving tasks.}
\label{fig:carla_lusr}
\end{figure*}

\subsection{Autonomous Driving In CARLA}
We further apply our approach in the autonomous driving task in CARLA simulator whose observation states are more complicated and thus increase the difficulty of RL training and generalization.

\subsubsection{LUSR Demonstration}
We also demonstrate the disentanglement of domain-specific embedding and domain-general embedding for images in CARLA simulator (see Figure \ref{fig:carla_lusr}). We first collect paired observation images from three tasks by placing the vehicle at the same starting point in the map and taking same actions in the driving. After collecting images, we extract their latent domain-general and domain-specific embeddings via a trained encoder and show their t-SNE plots in Figure \ref{fig:tsne1} and \ref{fig:tsne2}. It shows that domain-general embeddings from different tasks do have close similarities while domain-specific embeddings from different tasks are well clustered separately. The result demonstrates the disentanglement of domain-general and domain specific embeddings for images in CARLA driving tasks and thus supports the feasibility of LUSR in more challenging scenarios.

\subsubsection{Domain Adaptation Performance}
For domain adaptation in CARLA autonomous driving tasks (see Table \ref{tab: carla_result}), RL agents trained with LUSR is able to achieve zero-shot policy transfer without performance loss for both two target domains. It also achieves best scores in all domains compared with other approaches which shows the training efficiency of LUSR. 

Different from in CarRacing games, DARLA fails to adapt the trained policy in two target domains in CARLA autonomous driving tasks. This may be due to DARLA increasing the disentanglement of the latent embedding at the sacrifice of information accuracy. For complicated observation states like images in CARLA, it's much more difficult to achieve good disentanglement of each latent unit and the problem of information loss in the latent embedding is more serious. This causes the quality of the latent embedding in DARLA to be worse than LUSR and VAE-Embedding. This argument is supported by the results that higher $\beta$ in DARLA leads to worse RL training performance.    

VAE-Embedding achieves similar training performance in the source domain as LUSR while its adaptation performance in target domains is worse. 
Besides that, its driving behavior in target domains is very different from the behavior in the source domain. When adapting to target domains, the RL agent drives much slower and frequently reaches the time limit of each episode. As shown in table \ref{tab: carla_result}, compared to the result in the source domain, RL agents trained with VAE-Embedding receive lower scores while spending more time steps in each episode when driving in target domains.

Although CURL achieves comparable training performance in the source domain as other approaches, it completely fails to generalize to the two target domains in CARLA. To understand the reason, we conducted a tSNE analysis for CURL. The result reveals clusters based on domain labels. We believe the reason is that domain specific features are very useful in learning to assign low similarities for two states from different domains during CURL training and thus much domain specific information resides in the embedding of CURL. This prevents generalization across domains.

CycleGAN is also able to generalize to target domains well. However, it relies on pixel-wise input and the training efficiency is limited compared with other approaches that utilize internal latent state representation.



\section{Conclusion}
In this work, we propose to disentangle domain-general embedding and domain-specific embedding in the latent state representation of RL and theoretically formalize it in the scenario of domain adaptation. We propose LUSR which utilizes the domain-general latent embedding as state representation and prove its efficiency in two RL tasks with different visual complexity. As a result, our work enhances the applicability of Deep RL to real-world tasks that need both good domain adaptation performance and high training efficiency.

\newpage
\section{ Acknowledgments}
This work was supported by the Defense Advanced Research Projects Agency (DARPA) via Air Force Research Laboratory (AFRL) Contract No. FA8750-18-C-0103 (Lifelong Learning Machines: L2M). Authors are also thankful to computing resources provided by CHASE-CI under NSF Grant CNS-1730158.

\bigskip

\bibliography{refer.bib}

\end{document}